\documentclass[sigconf]{acmart}
\usepackage{caption}

\setcopyright{none}
\renewcommand\footnotetextcopyrightpermission[1]{}

\title{InfraPatch: Cross-Task Targeted Grayscale Patch Attacks on Infrared-Adapted Vision-Language Models}

\author{Chengyin Hu, Dingyi Lu, Jiaju Han, Xiang Chen, Weiwen Shi, Jiahuan Long, Yiwei Wei, Jiujiang Guo}

\begin{abstract}
Infrared vision-language models (IR-VLMs) have emerged as a promising paradigm
for multimodal perception under low-visibility conditions, yet their robustness
to targeted adversarial attacks remains poorly understood. Existing adversarial
patch methods mainly study RGB-based models or a single downstream task and do
not characterize whether localized perturbations can induce an intended
semantic target in IR-VLMs. We propose InfraPatch, a white-box, per-instance
framework for targeted digital grayscale patch attacks against IR-VLMs.
InfraPatch optimizes a compact single-channel patch within an approximately
$5\%$ local-area budget, combines proxy-guided placement with task-adaptive
semantic objectives, and induces target behaviors in image classification,
image captioning, and binary visual question answering. We evaluate ten
infrared-adapted model variants on 300 synthetic infrared-style images generated
by applying DiffV2IR to a fixed 30-category COCO subset, using clean-conditioned
targeted success criteria. InfraPatch achieves targeted attack success rates
from $86.00\%$ to $100\%$ across the ten variants. On CLIP and BLIP-2, proxy
location search improves success by $6.67$ and $10.33$ percentage points over
optimized random placement, respectively; LLaVA-1.5 remains saturated near
$100\%$ under both settings. Patch-area and objective ablations further expose
substantial differences in vulnerability across architectures and task
formats. These results show that small grayscale patches can inject chosen
target semantics across IR-VLM families under a controlled digital threat
model, motivating stronger robustness evaluation for infrared multimodal
systems.
\end{abstract}


\begin{document}

\maketitle

\section{Introduction}

Infrared sensing provides visual information in conditions where visible-light imagery can be degraded by insufficient illumination and other low-visibility factors \cite{hwang2015multispectral}. When coupled with vision-language models, infrared observations can be converted into categorical predictions, open-ended descriptions, or answers to task-specific questions. Such semantic outputs may inform a downstream perception pipeline or a human decision-maker rather than remaining an isolated model score. Consequently, a perturbation that steers an infrared vision-language model (IR-VLM) toward a chosen but unsupported semantic target represents a more specific risk than an untargeted reduction in accuracy.

Prior work has studied attacks on infrared classifiers and detectors, targeted attacks on RGB vision--language models, and illumination-transformation attacks that expose RGB-VLM sensitivity to lighting shifts \cite{edwards2020infrared,wei2023infraredpatch,wei2023unified,zhao2023evaluating,liu2025lighting}. Yet, to the best of our knowledge, no prior work has systematically tested whether a localized grayscale perturbation can inject one attacker-chosen semantic target across heterogeneous outputs of infrared-adapted vision--language models. In particular, a target-class flip in classification does not directly characterize target-phrase generation or answer manipulation in visual question answering (VQA). A unified cross-task formulation is needed to compare these behaviors without conflating different notions of attack success.

This cross-task setting presents both an optimization challenge and an evaluation challenge. For image classification, the model produces a fixed-dimensional class distribution, and a targeted success can be defined by the target becoming top-ranked with a positive margin over competing classes. For conditional text generation, the output is a free-form token sequence; the attack must optimize a target phrase and subsequently determine whether the generated text contains the intended object term under explicit normalization rules. For binary VQA, the relevant event is narrower: an image is valid only when the clean model explicitly answers ``no'' to a target-object question, and the attack succeeds only when the adversarial answer becomes ``yes''. These outputs require different differentiable objectives, valid-sample filters, and success parsers. Reusing a classification-only attack metric would therefore obscure the task-specific semantics and could overestimate success when the clean output already contains the target.

To study this problem, this paper introduces \emph{InfraPatch}, a white-box, per-instance framework for targeted digital grayscale patch attacks on IR-VLMs. InfraPatch restricts the perturbation to one square, single-channel patch with a nominal area budget of 5\% of the preprocessed image. The optimized grayscale values are replicated across the three model input channels and projected to the valid image range. The framework combines an early-exit proxy location search with task-adaptive semantic objectives. For each image, the search sequentially optimizes uniformly sampled legal placements under the complete attack objective. It immediately accepts the first candidate that satisfies the task-specific targeted-success criterion. The objective is adapted to target-class prediction, target-phrase generation, or a target VQA answer. The main contributions are:

\begin{itemize}
    \item To the best of our knowledge, the first systematic cross-task study of targeted adversarial patches for infrared-adapted VLMs, spanning classification, generation, and binary VQA, with success conditioned on clean outputs.
    \item A unified attack framework that couples early-exit proxy location search with task-adaptive semantic objectives under a compact, nominal 5\% grayscale-patch constraint.
    \item An evaluation design spanning ten model variants from the CLIP, BLIP, OpenFlamingo, and LLaVA families across three output formats, with task-specific success parsing and explicit exclusion of samples whose clean outputs already satisfy the target condition.
    \item Code-aligned comparisons with AdvIB, AdvIC, and HCB on one representative model per task, preserving each baseline's structured perturbation and derivative-free search while unifying the victim checkpoint, target, preprocessing, valid samples, nominal area budget, and final success parser.
\end{itemize}

\section{Related Work}

\subsection{Adversarial Patches and Targeted Attacks}

An adversarial patch confines optimized pixels to a local image region and
relaxes the small-norm constraint commonly imposed on image-wide examples.
Brown et al. introduced universal patches that remain robust under image
transformations and
drive an image classifier toward an attacker-chosen class
\cite{brown2017adversarialpatch}. Physical-world studies subsequently optimized
localized patterns under changes in viewpoint, distance, illumination, and
printing; for example, robust visual perturbations can induce targeted traffic-
sign misclassification after re-imaging \cite{eykholt2018robust}. These studies
establish that locality does not preclude strong targeted behavior, but their
success events are defined over a fixed classifier or detector output.

The optimization and evaluation of such attacks also build on broader
adversarial-robustness principles. Gradient-based attacks exploit the local
sensitivity of learned decision functions \cite{goodfellow2015explaining},
while strong optimization-based evaluations show that robustness conclusions
depend on whether the attack objective and stopping criterion faithfully test
the claimed property \cite{carlini2017evaluating}. Robust optimization instead
casts training as a worst-case inner maximization problem
\cite{madry2018towards}. For physical evaluation, expectation over
transformation (EOT) explicitly optimizes over a distribution of views,
scales, and other nuisance factors \cite{athalye2018synthesizing}.

Localized attacks have also been adapted from image classification to person
detection. Thys et al. optimized printable patches to suppress person
detections in surveillance imagery \cite{thys2019fooling}, and Xu et al.
modeled non-rigid deformation to realize an adversarial T-shirt under changing
poses and viewpoints \cite{xu2020tshirt}. Together with robust traffic-sign
perturbations, these studies demonstrate that a physically evaluated patch
requires transformation-aware optimization and re-imaging tests. InfraPatch
shares their localized threat surface but not their physical claim: its
single-channel constraint controls the digital input representation and should
not be interpreted as a thermal material model.

Infrared attacks face different realization constraints because thermal cameras
respond to emitted radiation rather than visible texture. Prior work demonstrated
adversarial examples on networks trained with simulated infrared imagery
\cite{edwards2020infrared}. Wei et al. optimized
the shape and placement of insulating patches to suppress
infrared pedestrian and vehicle detectors \cite{wei2023infraredpatch}, and later
optimized one shape-based patch against visible and infrared detectors
simultaneously \cite{wei2023unified}. Both methods target object detection in the
physical world. Score-based infrared attacks further exploit wearable hot--cold
structures. HCB jointly optimizes the shape and placement of grid-structured
thermal blocks \cite{wei2023hotcold}; AdvIB searches multiple rotatable infrared
blocks with differential evolution \cite{hu2024advib}; and AdvIC uses particle
swarm optimization to search two quadratic B\'{e}zier curves
\cite{hu2024advic}. These methods were originally designed as untargeted
disappearance attacks against infrared pedestrian detectors; here their
representations and searches are adapted to targeted IR-VLM objectives. Complementary
defense work uses thermal-radiation modeling to
guide adversarial training for infrared object detection
\cite{zhao2026infraredtraining}. InfraPatch instead studies a digital, single-channel intensity
patch and asks whether the same chosen semantic can be induced across
classification, conditional generation, and VQA; it therefore does not claim
the physical realizability established by thermal-material attacks.

\subsection{Robustness of Vision-Language Models}

Vision--language pre-training creates shared visual--textual representations,
which expose attack surfaces beyond conventional class logits; a recent survey
organizes these attacks by modality, objective, and model access
\cite{liu2025vlmattacksurvey}. Targeted visual
examples crafted on CLIP- or BLIP-like surrogates can transfer to generative
models and elicit attacker-specified responses \cite{zhao2023evaluating}, while
set-level cross-modal guidance improves transfer across pre-trained
vision--language models and downstream tasks \cite{lu2023sga}. At the generative
end of the spectrum, image-hijack attacks optimize visual inputs to make LLaVA
follow an adversary-selected behavior \cite{bailey2024imagehijacks}. Visual
adversarial examples and typographic visual prompts can also bypass multimodal
safety alignment \cite{qi2024visualjailbreak,gong2025figstep}. These
results demonstrate that manipulating the visual modality can alter linguistic
behavior, not merely decrease recognition accuracy.

Recent studies further examine adversarial VLM behavior in autonomous-driving
perception \cite{zhang2026autonomousvlm} and construct spatial--spectral
perturbations for physical-world large VLMs \cite{liu2026spatialspectral}.
Although these works broaden the operational settings of VLM attacks, they do
not define one clean-conditioned target semantic across infrared-adapted
classification, conditional generation, and binary VQA.

Robustness is also coupled to the shared visual backbone. Unsupervised
adversarial fine-tuning of a CLIP encoder has been shown to improve robustness
for zero-shot classification and for downstream large VLMs that reuse that
encoder \cite{schlarmann2024robustclip}. Multimodal defense can additionally
exploit one-to-many relationships between images and valid descriptions to
reduce dependence on a single paired target \cite{waseda2026multimodaldefense}.
Existing evaluations nevertheless
typically use image-wide norm-bounded perturbations, optimize a complete
adversarial image, or specialize the objective and success metric to one task.
InfraPatch complements them with a localized grayscale budget and
clean-conditioned, task-specific definitions of the same targeted semantic
event across three heterogeneous output spaces.

\subsection{Infrared Perception and Visible-to-Infrared Translation}

Thermal and visible sensing have long been treated as complementary modalities:
the KAIST multispectral benchmark, for instance, paired aligned color and
thermal video to study pedestrian detection across day and night conditions
\cite{hwang2015multispectral}. Modern vision--language systems broaden the
relevant output space. CLIP supports open-vocabulary classification through
contrastive image--text alignment \cite{radford2021clip}; BLIP-2 connects frozen
image encoders and language models through a lightweight query transformer
\cite{li2023blip2}; and InstructBLIP adapts this interface to instruction-
conditioned tasks \cite{dai2023instructblip}. OpenFlamingo provides an open
autoregressive framework for interleaved vision--language generation
\cite{awadalla2023openflamingo}, whereas the LLaVA line uses visual instruction
tuning and a compact vision--language connector for conversational and VQA
behavior \cite{liu2024llava15}.

Visible-to-infrared (V2IR) translation provides a way to construct synthetic
infrared-style inputs while retaining the semantic content of visible-image
benchmarks. DiffV2IR combines a Progressive Learning Module with a
Vision-Language Understanding Module to improve semantic awareness and
structure preservation during V2IR diffusion \cite{ran2025diffv2ir}. We use
DiffV2IR only as a data-generation tool and study the robustness of downstream
infrared-adapted VLMs. InfraPatch focuses on localized targeted perturbations
while retaining separate output parsers and valid-sample rules for
classification, generation, and binary VQA.

\section{Threat Model and Problem Formulation}

\label{sec:problem}

We study a frozen infrared-adapted vision--language model $f_{\theta}^{\tau}$ for
task $\tau\in\{\mathrm{cls},\mathrm{gen},\mathrm{vqa}\}$. For input
$x\in[0,1]^{3\times H\times W}$, prompt $q$, and fixed target semantic $t$
(``giraffe'' in all experiments), the evaluator applies a task-specific parser
$h_{\tau}$ to model outputs. The same parser is used for clean and adversarial
outputs.

\subsection{Digital Threat Model}

The attacker has white-box access to the frozen checkpoint, image processor,
prompt, target, and differentiable task loss. It may choose one square location
and one grayscale patch but may not alter model parameters, the prompt, pixels
outside the mask, or the output parser. The patch is clipped to $[0,1]$ and
replicated across the three input channels. We record its realized area after
preprocessing:
\begin{equation}
    \rho(x,m_{\ell})=\frac{\lVert m_{\ell}\rVert_0}{HW}=\frac{s^2}{HW}.
    \label{eq:area_ratio}
\end{equation}
The default budget is approximately $5\%$. This is a per-instance digital attack:
we do not claim universal transfer, EOT robustness, sensor re-imaging, or
physical thermal materials. The adapted AdvIB, AdvIC, and HCB baselines retain
derivative-free search but use the same victim, target, preprocessing, area
budget, and final parser as InfraPatch.

\subsection{Clean-Conditioned Target Semantics}

For classification, a sample is valid only when the clean top-1 prediction is
correct and non-target. Success requires the target to rank first with a
positive margin:
\begin{equation}
    \hat c(x_{\mathrm{adv}})=t,\qquad
    \pi_t(x_{\mathrm{adv}})-\max_{c\ne t}\pi_c(x_{\mathrm{adv}})>0.02.
    \label{eq:cls_success}
\end{equation}
For generation, a sample is valid when clean text does not contain the target.
Lowercasing, punctuation removal, singular/plural normalization, and a fixed COCO
synonym map define target hits in adversarial free-form outputs. For binary VQA,
the prompt is ``Is there a $t$ in the image? Answer only yes or no''. Valid
samples have clean output ``no''; success is an unambiguous transition to
``yes''. Outputs containing both tokens are rejected.

\subsection{Evaluation Metric}

Let $v_i^{\tau}$ denote clean validity and
$a_i^{\tau}=h_{\tau}(f_{\theta}^{\tau}(x_{i,\mathrm{adv}},q_i),t)$. We report
one cross-task primary metric:
\begin{equation}
    N_{\mathrm{valid}}^{\tau}=\sum_i v_i^{\tau},\qquad
    \operatorname{ASR}_{\mathrm{target}}^{\tau}
    =\frac{\sum_i v_i^{\tau}a_i^{\tau}}{N_{\mathrm{valid}}^{\tau}}.
    \label{eq:targeted_asr}
\end{equation}
Clean target hits enter the denominator only and are never counted as attack
success. We also record realized area, patch TV, and target margin when
meaningful in the corresponding output space.
\section{InfraPatch}

\label{sec:method}

\subsection{Method Overview}

For an infrared image $x$, a task prompt $q$, and a chosen target semantic $t$,
InfraPatch jointly selects a placement mask $m$ and optimizes a single-channel
patch $p$. The adversarial input is
\begin{equation}
    x_{\mathrm{adv}}=(1-m)\odot x+m\odot \operatorname{rep}_3\!\left(\Pi_{[0,1]}(p)\right),
    \label{eq:adv_input}
\end{equation}
where the single-channel mask is broadcast over the image channels,
$\operatorname{rep}_3(\cdot)$ replicates grayscale values three times, and
$\Pi_{[0,1]}(\cdot)$ projects them onto the valid image range. InfraPatch uses
a task-specific differentiable loss for classification, conditional text
generation, or binary VQA. It combines this term with total-variation (TV)
regularization to form the complete attack objective
\begin{equation}
    \mathcal{J}_{\mathrm{full}}(p,m;x,q,t)
    =\mathcal{L}_{\mathrm{task}}(p,m;x,q,t)
    +\lambda_{\mathrm{TV}}\mathcal{L}_{\mathrm{TV}}(p,m).
    \label{eq:full_objective}
\end{equation}
For classification, $\mathcal{L}_{\mathrm{task}}$ is the weighted combination
of target-class, target-similarity, competitor-suppression, and
original-class-suppression terms. For conditional generation and VQA, it is
the corresponding sequence objective defined below. The same
$\mathcal{J}_{\mathrm{full}}$ optimizes and ranks proxy candidates, aligning
placement selection with the continuation stage.

\begin{figure*}[t]
    \centering
    \includegraphics[width=0.92\textwidth]{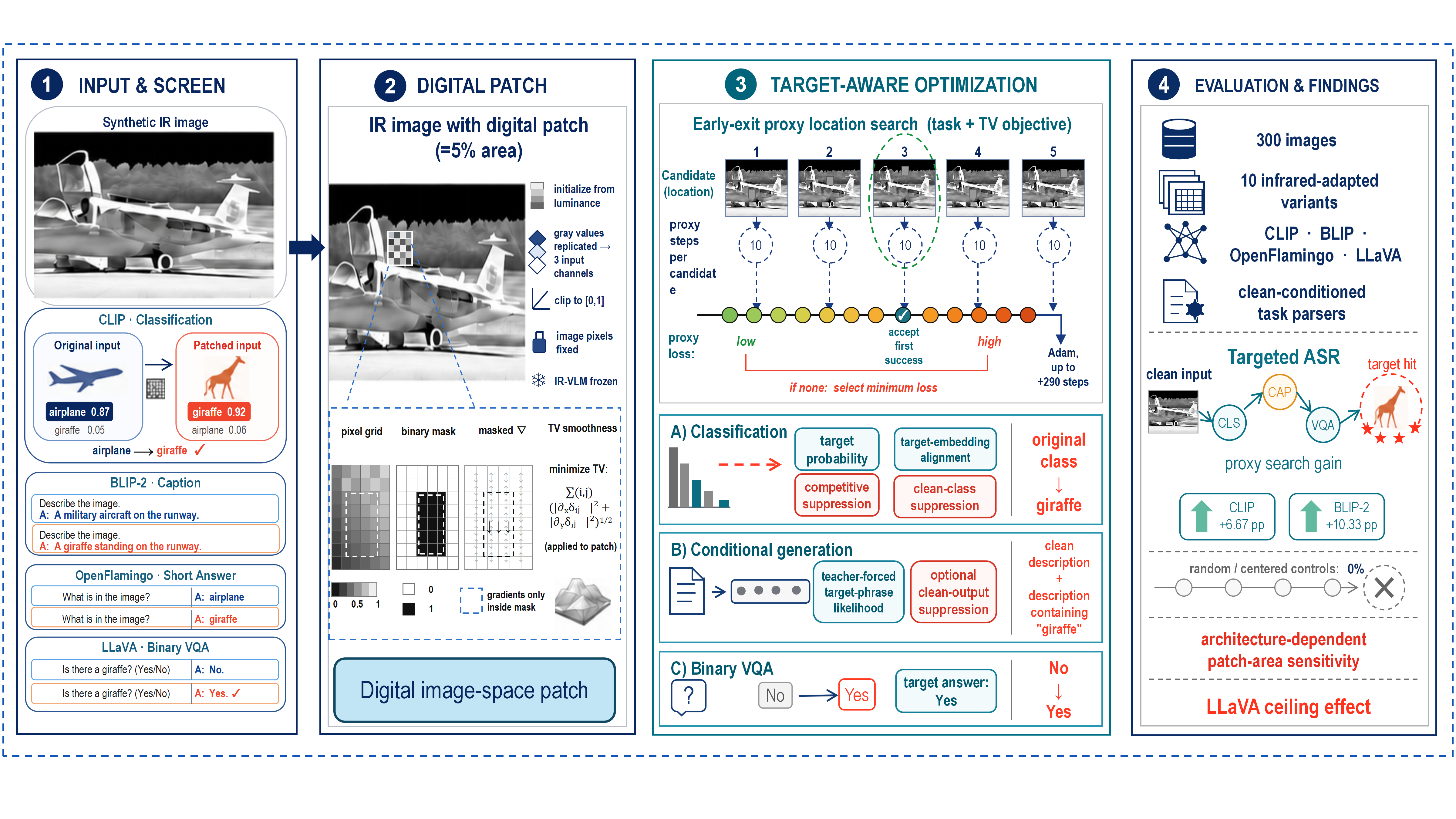}
    \caption{InfraPatch pipeline. Given a clean DiffV2IR-translated image and a
    fixed target semantic, the method first evaluates candidate patch locations
    through early-exit proxy search. It then refines the selected grayscale patch
    with a task-specific objective for classification, conditional generation,
    or binary VQA, and returns the adversarial output under the corresponding
    success rule.}
    \Description{A clean DiffV2IR-translated image, prompt, and fixed target enter
    a five-candidate early-exit location search. The selected location and
    grayscale patch are then refined with the objective and success rule for
    classification, conditional generation, or binary VQA.}
    \label{fig:infrapatch_overview}
\end{figure*}

Each candidate starts from the input luminance
\begin{equation}
    p^{(0)}=0.299x_R+0.587x_G+0.114x_B.
    \label{eq:gray_init}
\end{equation}
Thus, all locations receive the same deterministic grayscale initialization,
while gradients are retained only inside the selected mask.
Figure~\ref{fig:infrapatch_overview} summarizes the end-to-end pipeline: a
DiffV2IR-translated input, task prompt, and fixed target semantic enter an
early-exit proxy location search; the selected square mask and grayscale patch
are then refined under the task-specific objective in
Eq.~\eqref{eq:full_objective} and evaluated with the corresponding discrete
success parser.

\subsection{Target-Aware Patch Placement}

As shown in Figure~\ref{fig:infrapatch_overview}, the placement and continuation
stages share the same complete objective and success parser, so candidate
locations are ranked under the same semantic target that governs the final
attack. InfraPatch performs an early-exit proxy location search before the continuation
stage. It independently samples $K=5$ legal top-left coordinates with
replacement and evaluates them in sampled order. Each candidate receives a
fresh copy of Eq.~\eqref{eq:gray_init} and a newly initialized Adam optimizer.
It is then optimized with the complete objective for
$T_{\mathrm{proxy}}=10$ steps. For candidate $k$, the proxy score is the
lowest complete objective recorded along this short trajectory,
\begin{equation}
    s_k=\min_{0\leq j\leq T_{\mathrm{proxy}}}
    \mathcal{J}_{\mathrm{full}}(p_k^{(j)},m_k;x,q,t).
    \label{eq:proxy_score}
\end{equation}
After the tenth update, the method decodes the task output and applies the
discrete success rule from Section~\ref{sec:problem}.

When a candidate first satisfies the targeted-success criterion, InfraPatch
accepts its tenth-update patch and terminates the search; later candidates and
the continuation stage are skipped. If all candidates fail, it selects
$k^{\star}=\arg\min_k s_k$ and continues from that candidate's tenth-update
patch for at most $290$ additional steps using a newly initialized Adam
optimizer. The score in Eq.~\eqref{eq:proxy_score} may occur before the final
proxy iterate, but the implementation deliberately warm-starts continuation
from the final iterate rather than a best-loss checkpoint. Task-specific
decoding checks can terminate continuation early. Consequently, the selected
placement receives at most $300$ updates, while the worst-case per-image cost
is $5\times10+290=340$ gradient updates. Throughout the paper, ``proxy
objective'' denotes the complete objective in Eq.~\eqref{eq:full_objective},
not only its target term.

\subsection{Task-Adaptive Semantic Objectives}

\subsubsection{Classification Objective}

For the CLIP-family models, let $z$ be the normalized image embedding and
$e_c$ the normalized embedding for class $c$, obtained by averaging the shared
prompt-template ensemble. With scaled logit
$g_c=\exp(\gamma)z^{\top}e_c$, target class $t$, current strongest non-target
competitor $r$, and clean predicted class $c_0$, the implemented loss is
\begin{align}
    \mathcal{L}_{\mathrm{cls}}
    ={}&\lambda_{\mathrm{ce}}[-\log\operatorname{softmax}(g)_t]
    +\lambda_{\mathrm{cos}}(1-z^{\top}e_t) \notag\\
    &+\lambda_{\mathrm{sup}}\max(z^{\top}e_r,0)
    +\lambda_{\mathrm{orig}}\operatorname{softmax}(g)_{c_0}.
    \label{eq:cls_loss}
\end{align}
The four default weights are one. The terms respectively increase the target
probability, align the image with the target text embedding, suppress the
strongest competitor, and reduce persistence of the clean prediction. The
optimization objective is differentiable, whereas success is determined by
the decoded top-1 label and margin in Eq.~\eqref{eq:cls_success}.

\subsubsection{Conditional-Generation Objective}

Let $y_t=(y_{t,1},\ldots,y_{t,L})$ be a short target phrase and let $y_0$ be
the clean generated text.  The conditional-generation objective uses
teacher-forced negative log likelihood (NLL):
\begin{equation}
    \mathcal{L}_{\mathrm{gen}}
    =\operatorname{cap}_{B}\!\left[\operatorname{NLL}(y_t\mid x_{\mathrm{adv}},q)\right]
    -\lambda_{\mathrm{clean}}
     \operatorname{cap}_{B}\!\left[\operatorname{NLL}(y_0\mid x_{\mathrm{adv}},q)\right],
    \label{eq:gen_loss}
\end{equation}
where $\operatorname{cap}_{B}(u)=\min(u,B)$. BLIP-2, BLIP-2-ViT-L, and
InstructBLIP use the clean-output suppression term with
$\lambda_{\mathrm{clean}}=0.30$ and $B=12$. OpenFlamingo uses the target-NLL
term with $\lambda_{\mathrm{clean}}=0$ and $B=100$. The default target phrase
is ``a giraffe''. Candidate and final success are always determined from
free-running generation, rather than from the teacher-forced loss alone.

\subsubsection{Visual Question Answering Objective}

For LLaVA, the prompt tokens are masked from the labels and the target sequence
contains ``Yes'' followed by the end-of-sequence token. The objective is
\begin{equation}
    \mathcal{L}_{\mathrm{vqa}}
    =-\frac{1}{L_a}\sum_{j=1}^{L_a}
      \log P_{\theta}(a_{t,j}\mid a_{t,<j},x_{\mathrm{adv}},q).
    \label{eq:vqa_loss}
\end{equation}
LLaVA-1.6 caps this NLL at $12$ for optimization, whereas the implemented
LLaVA-1.5 variant uses the uncapped NLL. In both cases, a generated and
unambiguous ``yes'' is required for success; a low teacher-forced NLL alone is
not counted.

\subsection{Patch Optimization and Constraints}

Having defined the task-specific terms in $\mathcal{L}_{\mathrm{task}}$, we
next specify how the patch parameters are updated under the shared constraints.
Throughout optimization, only the patch $p$ is optimized; all model and
text-embedding parameters remain frozen. The proxy and continuation stages use
Adam with initial learning rate $0.03$.
The continuation stage applies cosine annealing to $5\%$ of that rate.  During
each forward pass, the rendered patch is clipped to $[0,1]$, replicated across
channels, and composited through $m$.  Gradients outside the mask are set to
zero.

The total-variation term penalizes only adjacent pixels that are both inside
the patch, excluding the patch--background boundary.  With horizontal and
vertical internal edge sets $\mathcal{E}_h$ and $\mathcal{E}_v$, it is
\begin{equation}
    \mathcal{L}_{\mathrm{TV}}
    =\frac{1}{|\mathcal{E}_h|}\sum_{(u,v)\in\mathcal{E}_h}|p_u-p_v|
    +\frac{1}{|\mathcal{E}_v|}\sum_{(u,v)\in\mathcal{E}_v}|p_u-p_v|,
    \label{eq:tv_loss}
\end{equation}
with empty denominators clamped to one in code.  The default coefficient is
$\lambda_{\mathrm{TV}}=10^{-3}$.  This term controls local smoothness but is
not treated as evidence of physical realizability.

\section{Experimental Setup}

\label{sec:experiments}

We evaluate ten infrared-adapted variants from four model families on three
output formats. Classification uses CLIP, OpenCLIP \cite{cherti2023openclip},
MetaCLIP \cite{xu2024metaclip}, and EVA-CLIP \cite{sun2023evaclip};
captioning uses BLIP-2, BLIP-2-ViT-L, InstructBLIP, and OpenFlamingo; binary VQA
uses LLaVA-1.5 and LLaVA-1.6. All checkpoints start from natural-image
pre-training, receive the same LoRA adaptation recipe \cite{hu2022lora} on
synthetic infrared-style training data, and remain frozen during attack
optimization. Classification adaptation uses an InfoNCE objective
\cite{oord2018cpc}, whereas the generation variants use next-token prediction.

\subsection{Data and Preprocessing}

The evaluation set is a fixed 300-image, 30-category MS COCO subset curated by
Liu et al.\ for their cross-task study of VLM robustness to illumination
transformations \cite{lin2014coco,liu2025lighting}, with ten images per
category. We apply DiffV2IR \cite{ran2025diffv2ir} to obtain infrared-style
digital inputs;
these are image-to-image synthesis results, not sensor measurements. The target
category ``giraffe'' is excluded from directory labels. Each task uses its native
processor and prompt: ``Describe the image'' for BLIP-family captioning,
``What is in the image?'' for OpenFlamingo short answers, and the strict yes/no
question defined in Section~\ref{sec:problem} for LLaVA. Fixed-resolution
pipelines use $50\times50$ patches at $224^2$ and LLaVA-1.5 uses $75\times75$ at
$336^2$, both realizing approximately $4.98\%$ area. LLaVA-1.6 retains its
any-resolution path and records the ratio per image.

\subsection{Baselines and Controls}

We adapt AdvIB, AdvIC, and HCB with the same victim, target, clean-valid set,
preprocessing, nominal area budget, and discrete success parser. Their defining
block, curve, and grid representations and derivative-free DE/PSO searches are
preserved; only the detector-disappearance fitness is replaced by a task-targeted
fitness. The comparison is area-controlled rather than query-matched.

For component analysis we use CLIP, BLIP-2, and LLaVA-1.5 as representatives.
Controls compare full InfraPatch with optimized random placement, equal-area
unoptimized random grayscale, and equal-area unoptimized center patches. Area
settings near $3\%$, $5\%$, and $7\%$ and one-factor objective/TV ablations use
the same image pool, target, initialization, budget, and parser. Full settings
and search budgets are listed in Appendix~\ref{app:impl_details} and
Appendix~\ref{app:baseline_details}.

\subsection{Evaluation Protocol}

We report $N_{\mathrm{valid}}$, successful attacks, targeted ASR, and two-sided
$95\%$ Wilson intervals from the displayed counts. InfraPatch uses Adam with
learning rate $0.03$, five proxy candidates, ten proxy updates per candidate,
and at most 290 continuation updates; it is purely digital and uses no EOT.
All comparisons use the clean-conditioned definitions in Section~\ref{sec:problem}.

\section{Results and Analysis}

\subsection{Main Results across Models and Tasks}

Table~\ref{tab:main_results} gives the primary clean-conditioned outcome before
we interpret task-specific patterns. It reports the valid denominator explicitly,
so a perfect score is not confused with a large but filtered sample.

\begin{table*}[t]
    \caption{Cross-model targeted attack results under the clean-conditioned
    protocol. Confidence intervals are two-sided $95\%$ Wilson intervals from
    the displayed counts.}
    \label{tab:main_results}
    \centering
    \small
    \begin{tabular}{lllrrrr}
        \toprule
        Family & Variant & Output / task & $N_{\mathrm{valid}}$ & Success & ASR (\%) & 95\% CI \\
        \midrule
        CLIP & CLIP ViT-L/14 & Classification & 165 & 157 & 95.15 & [90.73, 97.52] \\
        CLIP & OpenCLIP ViT-B/16 & Classification & 132 & 132 & \textbf{100.00} & [97.17, 100.00] \\
        CLIP & MetaCLIP ViT-L/14 & Classification & 173 & 158 & 91.33 & [86.19, 94.68] \\
        CLIP & EVA-CLIP EVA-G/14+ & Classification & 180 & 167 & 92.78 & [88.04, 95.73] \\
        BLIP & BLIP-2 Flan-T5-XL & Caption generation & 300 & 258 & 86.00 & [81.62, 89.47] \\
        BLIP & BLIP-2 ViT-L & Caption generation & 300 & 289 & 96.33 & [93.55, 97.94] \\
        BLIP & InstructBLIP Flan-T5-XL & Caption generation & 300 & 262 & 87.33 & [83.09, 90.63] \\
        OpenFlamingo & ViT-L/14 + MPT-1B & Short answer & 299 & 283 & 94.65 & [91.49, 96.68] \\
        LLaVA & LLaVA-1.5-7B & Binary VQA & 286 & 286 & \textbf{100.00} & [98.67, 100.00] \\
        LLaVA & LLaVA-1.6-Mistral-7B & Binary VQA & 300 & 300 & \textbf{100.00} & [98.74, 100.00] \\
        \bottomrule
    \end{tabular}
\end{table*}

\begin{figure*}[t]
    \centering
    \includegraphics[width=\textwidth]{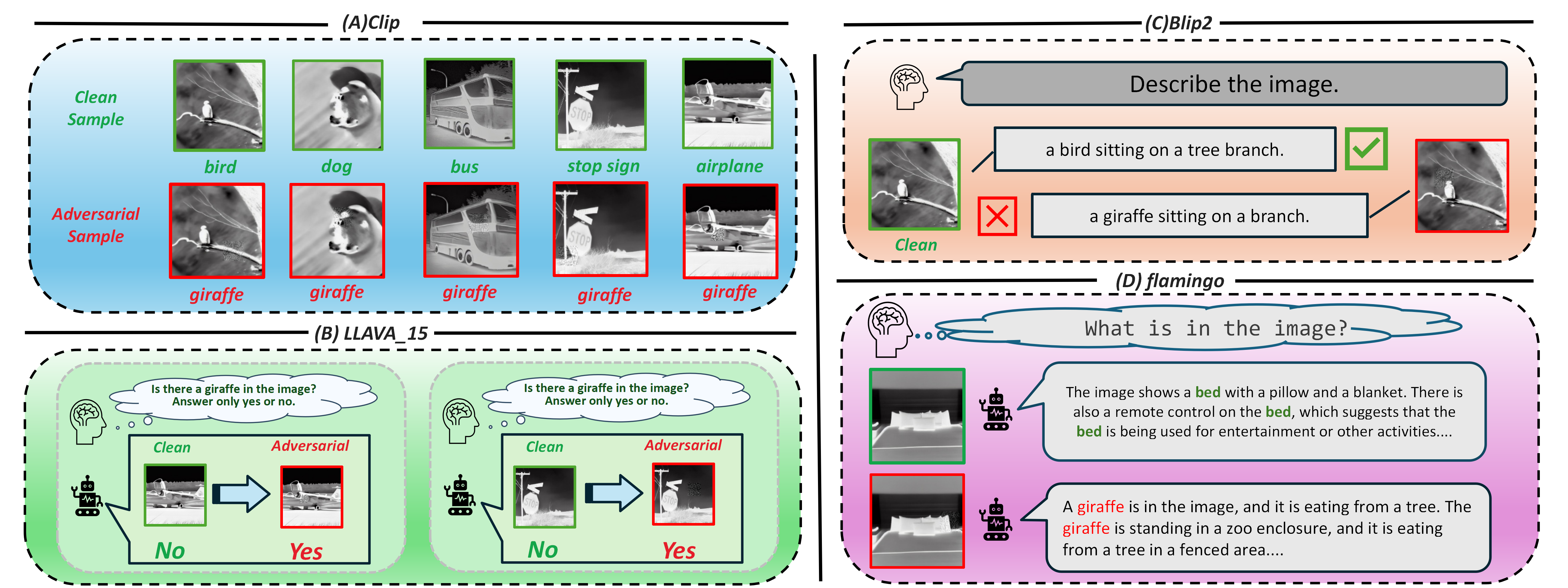}
    \caption{Qualitative cross-task examples under InfraPatch on
    DiffV2IR-translated inputs.
    \textbf{(A) Image classification (OpenAI CLIP):} five categories (bird, dog,
    bus, stop sign, airplane); clean predictions match the source label, while
    adversarial inputs are classified as \textit{giraffe}.
    \textbf{(B) Binary visual question answering (LLaVA-1.5):} prompt ``Is there
    a giraffe in the image? Answer only yes or no''.
    \textbf{(C) Image captioning (BLIP-2):} airplane sample with prompt
    ``Describe the image''.
    \textbf{(D) Image captioning (OpenFlamingo):} bed sample. Green denotes clean
    outputs; red denotes adversarial outputs or successful target injection.}
    \Description{Four-panel figure showing InfraPatch qualitative cross-task
    examples on image classification, binary visual question answering, and
    image captioning across DiffV2IR-translated inputs.}
    \label{fig:cross_task_qualitative}
\end{figure*}

Figure~\ref{fig:cross_task_qualitative} previews representative clean-conditioned
successes across classification, binary VQA, and captioning under the same
nominal $5\%$ grayscale budget and evaluation rules. Panel~(A) shows that a
single localized patch can flip top-1 classification on five non-target
categories while leaving the surrounding scene otherwise plausible under
DiffV2IR translation. Panel~(B) demonstrates the narrower VQA event: the clean
model answers ``no'', and the adversarial input elicits an unambiguous ``yes''
under the same parser. Panels~(C)--(D) further show that the target semantic
survives free-form generation, appearing in BLIP-2 captions and OpenFlamingo
short answers rather than only in class logits.

InfraPatch achieved high clean-conditioned targeted success across all ten
IR-VLM variants (Table~\ref{tab:main_results}). Targeted ASR ranged from
$86.00\%$ on BLIP-2 to $100\%$ on OpenCLIP and both LLaVA variants. Among the
four classifiers, CLIP, MetaCLIP, and EVA-CLIP reached $95.15\%$, $91.33\%$,
and $92.78\%$, respectively; OpenCLIP achieved 132 of 132 successes ($100\%$).
Thus, even under the strict clean-conditioned protocol,
InfraPatch remains highly effective on contrastive classifiers, with three
variants above $91\%$ and one reaching perfect success.

The conditional-generation results also generalized beyond one architecture.
BLIP-2-ViT-L reached $96.33\%$, OpenFlamingo reached $94.65\%$ under the strict
$283/299$ clean-conditioned count, and InstructBLIP and BLIP-2 reached
$87.33\%$ and $86.00\%$. Both binary-VQA variants reached $100\%$, corresponding
to $286/286$ valid LLaVA-1.5 inputs and $300/300$ LLaVA-1.6 inputs. As shown in
Figure~\ref{fig:cross_task_qualitative}, the target event was not confined to
class logits: it also appeared in free-form generation and answer transitions
under their task-specific parsers.

\subsection{Comparison with Adapted Structured Infrared Attacks}

InfraPatch achieved the highest targeted ASR on all three representative
model--task pairs in Table~\ref{tab:adapted_baselines}. On CLIP classification,
it reached $95.15\%$, versus $9.70\%$ for AdvIC-Adapted; the corresponding
BLIP-2 rates were $86.00\%$ and $7.00\%$, and the LLaVA-1.5 rates were
$100.00\%$ and $14.69\%$. AdvIC was the strongest adapted baseline on every
task, while all three structured attacks remained between $1.33\%$ and
$14.69\%$. Under the implemented budgets, detector-oriented block, curve, and
grid representations therefore transfer weakly to targeted IR-VLM semantics.
Because search budgets differ, this comparison evaluates each complete
representation--optimizer combination rather than isolating geometry or access
cost.

\begin{table}[t]
    \caption{Clean-conditioned targeted ASR (\%) for InfraPatch and adapted
    structured infrared attacks on one representative model per task.}
    \label{tab:adapted_baselines}
    \centering
    \small
    \resizebox{\columnwidth}{!}{%
    \begin{tabular}{lrrrrr}
        \toprule
        Model / task & $N_{\mathrm{valid}}$ & HCB & AdvIB & AdvIC & InfraPatch \\
        \midrule
        CLIP / Classification & 165 & 4.24 & 7.88 & 9.70 & 95.15 \\
        BLIP-2 / Captioning & 300 & 1.33 & 3.67 & 7.00 & 86.00 \\
        LLaVA-1.5 / Binary VQA & 286 & 2.80 & 6.99 & 14.69 & 100.00 \\
        \bottomrule
    \end{tabular}}
\end{table}

\subsection{Comparison with Internal Controls}

Semantic optimization was necessary for the observed target hits. Equal-area
unoptimized random and center grayscale patches produced $0\%$ targeted ASR on
all three representative variants; this means they did not induce the chosen
target, not that model outputs were unchanged.

\begin{table}[t]
    \caption{Location-selection and optimization controls. Cells report targeted
    ASR (\%); optimized random placement is mean $\pm$ standard deviation over
    three seeds.}
    \label{tab:location_controls}
    \centering
    \small
    \begin{tabular}{lccc}
        \toprule
        Setting & CLIP & BLIP-2 & LLaVA-1.5 \\
        \midrule
        Full InfraPatch & 95.15 & 86.00 & 100.00 \\
        Random loc., optimized & 88.48$\pm$1.60 & 75.67$\pm$0.88 & 99.88$\pm$0.20 \\
        Random gray, unoptimized & 0.00 & 0.00 & 0.00 \\
        Center gray, unoptimized & 0.00 & 0.00 & 0.00 \\
        \bottomrule
    \end{tabular}
\end{table}

\begin{figure}[t]
    \centering
    \includegraphics[width=\columnwidth]{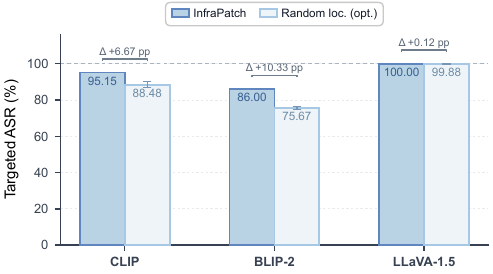}
    \caption{Targeted ASR under full InfraPatch and optimized random placement.
    Error bars denote sample standard deviation across three seeds.}
    \Description{A grouped bar chart compares full InfraPatch and optimized
    random placement on CLIP, BLIP-2, and LLaVA-1.5.}
    \label{fig:placement_controls}
\end{figure}

Figure~\ref{fig:placement_controls} compares full InfraPatch with optimized
random placement. A single optimized random location was
already strong, but proxy search increased mean ASR from $88.48\%$ to
$95.15\%$ on CLIP and from $75.67\%$ to $86.00\%$ on BLIP-2, gains of $6.67$
and $10.33$ percentage points. On LLaVA-1.5, random placement averaged
$99.88\%$ and InfraPatch reached $100\%$, leaving little headroom for location
selection.

This comparison is not strictly compute matched. The single-location baseline
receives at most 300 updates, whereas InfraPatch can spend $5\times10+290=340$
updates when no proxy candidate exits early. The observed difference therefore
quantifies the complete proxy-search procedure, including its additional
location evaluations, rather than an isolated placement effect at identical
gradient cost.

\subsection{Ablation Studies}

The ablations expose task-dependent contributions (Table~\ref{tab:ablations}).
\begin{table}[t]
    \caption{Patch-area and objective ablations. Cells report targeted ASR (\%);
    dashes denote configurations not applicable to a task.}
    \label{tab:ablations}
    \centering
    \small
    \begin{tabular}{lccc}
        \toprule
        Configuration & CLIP & BLIP-2 & LLaVA-1.5 \\
        \midrule
        Full, $\rho\approx5\%$ & 95.15 & 86.00 & 100.00 \\
        $\rho\approx3\%$ & 92.73 & 76.67 & 100.00 \\
        $\rho\approx7\%$ & 96.97 & 89.67 & 100.00 \\
        $\lambda_{\mathrm{TV}}=0$ & 94.55 & 82.00 & 100.00 \\
        $\lambda_{\mathrm{cos}}=0$ & 92.73 & -- & -- \\
        $\lambda_{\mathrm{sup}}=0$ & 96.36 & -- & -- \\
        $\lambda_{\mathrm{orig}}=0$ & 97.58 & -- & -- \\
        $\lambda_{\mathrm{clean}}=0$ & -- & 86.33 & -- \\
        \bottomrule
    \end{tabular}
\end{table}

Removing target-embedding alignment reduced CLIP ASR by $2.42$ points, from
$95.15\%$ to $92.73\%$. Removing TV reduced CLIP by only $0.60$ points but
reduced BLIP-2 by $4.00$ points, while LLaVA-1.5 remained saturated. In
contrast, removing rival suppression or clean-class suppression increased CLIP
ASR to $96.36\%$ and $97.58\%$, and disabling BLIP-2 clean-output suppression
changed ASR from $86.00\%$ to $86.33\%$. These results do not support claiming
that every loss term improves success; under the current target and benchmark,
some terms appear redundant for ASR but may affect other properties, such as
optimization dynamics or patch smoothness.

\subsection{Patch-Area Sensitivity}

\begin{figure}[t]
    \centering
    \includegraphics[width=\columnwidth]{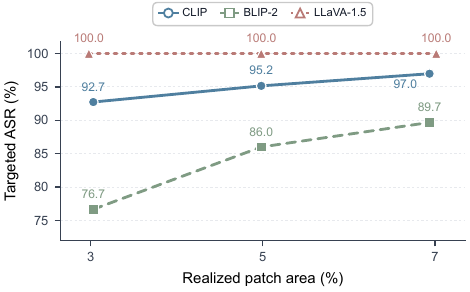}
    \caption{Targeted ASR as a function of realized patch area for the
    representative classification, conditional-generation, and binary-VQA
    variants.}
    \Description{A line plot shows targeted attack success rate at approximately
    three, five, and seven percent patch area for CLIP, BLIP-2, and LLaVA-1.5.}
    \label{fig:area_asr}
\end{figure}

Increasing patch area had the clearest effect on BLIP-2
(Table~\ref{tab:ablations}; Figure~\ref{fig:area_asr}). Moving from approximately $3\%$ to $5\%$ and
$7\%$ raised its ASR from $76.67\%$ to $86.00\%$ and $89.67\%$. CLIP rose more
gradually, from $92.73\%$ to $95.15\%$ and $96.97\%$. The corresponding
realized fixed-resolution ratios were $3.03\%$, $4.98\%$, and $6.94\%$ for
$224\times224$ inputs, and $2.98\%$, $4.98\%$, and $7.01\%$ for
$336\times336$ inputs. LLaVA-1.5 remained at $100\%$ throughout, so these data
identify a ceiling effect rather than area invariance. Because no perceptual
study or thermal-sensor measurement was archived, patch area is treated as a
budget variable and not as a direct measure of visual stealth.

\subsection{Robustness Checks and Failure Cases}

Failure rates differed across families: BLIP-2 failed on 42 of 300 valid samples,
InstructBLIP on 38, BLIP-2-ViT-L on 11, MetaCLIP on 15 of 173, and EVA-CLIP on
13 of 180. These counts caution against generalizing perfect OpenCLIP and LLaVA
results to all IR-VLMs. Because aggregate summaries lack per-sample traces, we
report failures without assigning semantic or spatial causes; clean-valid
denominators also differ across classification, captioning, and VQA.

\subsection{Cross-Task Interpretation and Boundary Analysis}

The three task families differ in output freedom. Classification
has a closed vocabulary but requires both a top-1 flip and a positive target
margin; clean-valid denominators (132--180) reflect clean recognition variation,
not attack budget. Within those filtered sets, all four classifiers exceed
$91\%$ targeted ASR. A relatively low clean-valid count with high conditional
ASR shows that the patch is effective once a sample enters the evaluation set,
not that it repairs clean recognition errors.

Generation has the most semantic freedom yet uses the strictest clean-
conditioned lexical event. BLIP-2, BLIP-2-ViT-L, and InstructBLIP span
$86.00\%$--$96.33\%$, while OpenFlamingo reaches $283/299$ after excluding one
clean target hit. Values are based on free-running decoding, not only
teacher-forced loss; the lexical parser remains the operational criterion.

Binary VQA restricts outputs to unambiguous ``yes'' or ``no''. Both LLaVA
variants reach $100\%$ on clean-valid sets, demonstrating reliable no-to-yes
transitions under the strict prompt but not open-ended transfer. The contrast
between perfect VQA and lower captioning ASR illustrates why one class-flip
metric would be insufficient for cross-task robustness.

The placement control separates semantic optimization from location selection.
Unoptimized grayscale controls produce no target hits, so success requires
learned patch values rather than an equal-area occluder alone. Optimized random
placement is already effective, but proxy search adds $6.67$ and $10.33$
percentage points on CLIP and BLIP-2. The gain is nearly absent on saturated LLaVA-1.5. With at most 300 versus
$5\times10+290=340$ updates, this supports the complete procedure, not a
compute-matched location estimate.

The ablations provide a complementary explanation. Target-embedding alignment
is the only removed CLIP term that clearly lowers ASR, whereas rival and clean-
class suppression are redundant for this target and sample pool. TV regulariza-
tion matters more for BLIP-2 than CLIP. Area sensitivity follows the same pattern:
BLIP-2 gains most from increasing the patch, CLIP changes more gradually, and
LLaVA remains saturated from approximately $3\%$ to $7\%$---architecture-
dependent trends, not evidence that larger patches are universally more effective
or visually stealthy.

Together, the results support a bounded claim. Under one synthetic DiffV2IR
distribution, one target semantic, a white-box digital threat model, and the
reported clean-conditioned parsers, localized grayscale injection is effective
across heterogeneous IR-VLM outputs. The same evidence does not establish
physical realizability, black-box transfer, universal patches, or robustness to
sensor noise. Those distinctions govern interpretation of the high ASR values
and motivate the limitations that follow.

The clean-conditioned denominator also shapes cross-model comparison. CLIP
clean-valid counts range from 132 to 180 because directory-label and top-1
agreement is required; captioning retains almost all 300 images, except one
clean target hit in OpenFlamingo, whereas VQA retains only images whose clean
answer is an unambiguous ``no''. Consequently, targeted ASR is a conditional
vulnerability measure rather than an end-to-end deployment probability;
reporting $N_{\mathrm{valid}}$ keeps comparisons auditable.

Parser design separates optimization from success. Classification success is
checked from the decoded top-1 label and margin; generation is checked from
normalized free-running text; and VQA rejects outputs containing both ``yes'' and
``no''. Teacher-forced target likelihood guides BLIP-family updates, but it
cannot by itself establish that the target phrase was generated. The qualitative
panels illustrate the same discrete events counted in Table~\ref{tab:main_results}.

Finally, baseline and ablation evidence are controlled comparisons under the
same victim checkpoint, target, processor, valid set, and area budget. Structured
baselines remain derivative-free, while InfraPatch uses gradients and can evaluate
several proxy locations. This design isolates whether detector-era geometries
transfer to a semantic target, not optimizer query cost. Each ablation row changes
one term at a time; the resulting differences support mechanism-level discussion,
but they should not be read as universal rankings of attack algorithms outside
this protocol.

\section{Limitations, Responsible Use, and Conclusion}

\subsection{Limitations}

The study is bounded to white-box, per-instance digital attacks on one
synthetic DiffV2IR-translated COCO set and one target semantic. It does not
establish universal or black-box transfer, sensor robustness, or physical
realizability; images are not re-imaged after printing or sensor noise. The
structured baseline comparison covers one model per task and is neither query
nor compute matched. The 300-image, 30-category set does not measure target or
long-tail variation, and the shared adaptation recipe cannot separate
pre-training from LoRA-induced vulnerability.

\subsection{Responsible Use}

InfraPatch is intended for authorized digital robustness evaluation of infrared
multimodal systems, not covert physical deployment. Artifact release should
include provenance, fixed parsers, model cards, and mitigations such as
patch-aware augmentation and output-consistency checks; safety-critical tests
should be isolated and authorized.

\subsection{Conclusion}

This paper introduced InfraPatch, a proxy-guided digital grayscale patch attack
for infrared-adapted vision-language models with heterogeneous output formats. Under a
clean-conditioned protocol, the attack reached $86.00\%$--$100\%$ targeted ASR
across ten infrared-adapted variants, and proxy location search improved CLIP
and BLIP-2 over optimized random placement. The area and objective ablations
showed that vulnerability and component utility vary by architecture, while
LLaVA exhibited a pronounced ceiling effect. Within its synthetic, digital,
white-box scope, the adapted AdvIB, AdvIC, and HCB attacks achieved only
$1.33\%$--$14.69\%$ ASR on the three representative tasks. This indicates that,
under the implemented budgets, detector-era infrared block, curve, and grid
structures transfer only weakly to targeted IR-VLM semantics, while localized
target-semantic injection remains an IR-VLM robustness concern.

\bibliographystyle{ACM-Reference-Format}
\bibliography{references}

\clearpage
\appendix

\section{Additional Implementation Details}
\label{app:impl_details}

\begin{table}[h]
    \caption{Prompts, target semantics, and supplementary attack settings.}
    \label{tab:impl_details}
    \centering
    \small
    \begin{tabular}{@{}p{0.34\columnwidth}p{0.58\columnwidth}@{}}
        \toprule
        Setting & Value \\
        \midrule
        Default target & ``giraffe'' \\
        BLIP-family prompt & ``Describe the image'' \\
        OpenFlamingo prompt & ``What is in the image?'' \\
        LLaVA prompt & ``Is there a giraffe in the image? Answer only yes or no'' \\
        Generation target phrase & ``a giraffe'' \\
        VQA target sequence & ``Yes'' \\
        Patch size ($224^2$ / $336^2$) & $50{\times}50$ / $75{\times}75$ \\
        Continuation LR schedule & cosine annealing to $5\%$ of initial rate \\
        TV coefficient & $10^{-3}$ \\
        Continuation optimizer & new Adam (no proxy-state carry-over) \\
        Decode cadence & BLIP/OpenFlamingo: every $100$ updates; LLaVA: every $50$ \\
        Software stack & PyTorch 2.0.1+cu117, CUDA 11.7, Transformers 4.46.3 \\
        \bottomrule
    \end{tabular}
\end{table}

\section{Adapted Structured Infrared Baseline Details}
\label{app:baseline_details}

The AdvIB adaptation uses seven rotatable square cold blocks at grayscale $0$.
The block side is derived from $0.05HW/7$, and differential evolution searches
the 21 normalized position-and-angle parameters with population 100, ten
generations, mutation factor $0.5$, crossover probability $0.6$, seed 42, and
early stopping, for at most 1,100 fitness evaluations. AdvIC-Adapted renders
two black quadratic B\'{e}zier curves specified by 12 normalized control-point
parameters. Stroke width is set from their rasterized length and the nominal
$5\%$ pixel budget; PSO uses 50 particles, ten iterations, seed 42, inertia
$0.9$, and cognitive/social coefficients $1.6/1.4$, for at most 500
evaluations. The instance-level HCB adaptation shares one thresholded $3\times3$ binary
state matrix across four optimized placements. Active cells use grayscale
$0.2$, their size is derived from the nominal budget, and PSO uses 100
particles and three iterations with the same seed and coefficients, for at
most 300 evaluations.

The derivative-free optimizers minimize continuous task fitness while final
ASR uses the same discrete success rules as InfraPatch
(Section~\ref{sec:problem}); continuous fitness definitions are provided in the
released code. All coordinates lie on the legal preprocessed canvas. Overlap or
clipping can make realized union area smaller than the nominal budget. Each run
retains the best fitness, parameters, realized mask, query count, decoded output,
and success status, including failures.

\end{document}